# Engineering deep learning methods on automatic detection of damage in infrastructure due to extreme events

SAGE

**Yongsheng Bai, Bing Zha, Halil Sezen and Alper Yilmaz**

## Abstract

This paper presents a few comprehensive experimental studies for automated Structural Damage Detection (SDD) in extreme events using deep learning methods for processing 2D images. In the first study, a 152-layer Residual network (ResNet) is utilized to classify multiple classes in eight SDD tasks, which include identification of scene levels, damage levels, material types, etc. The proposed ResNet achieved high accuracy for each task while the positions of the damage are not identifiable. In the second study, the existing ResNet and a segmentation network (U-Net) are combined into a new pipeline, cascaded networks, for categorizing and locating structural damage. The results show that the accuracy of damage detection is significantly improved compared to only using a segmentation network. In the third and fourth studies, end-to-end networks are developed and tested as a new solution to directly detect cracks and spalling in the image collections of recent large earthquakes. One of the proposed networks can achieve an accuracy above 67.6% for all tested images at various scales and resolutions, and shows its robustness for these human-free detection tasks. As a preliminary field study, we applied the proposed method to detect damage in a concrete structure that was tested to study its progressive collapse performance. The experiments indicate that these solutions for automatic detection of structural damage using deep learning methods are feasible and promising. The training datasets and codes will be made available for the public upon the publication of this paper.

## Introduction

Artificial Intelligence (AI) began as an academic research subject in 1950s, and currently the commercialization and potential applications of AI are being pushed for almost all industries. Machine learning, a sub-subject of AI technology, is a vital discipline developing various algorithms for learning from data, identifying patterns and making decisions without human intervention. As a subcategory of machine learning, deep learning provides state of the art results to problems that are initially considered to be intuitively solved by humans. Deep learning models learn from experiences and evolve through training and testing, and they are particularly effective at learning complicated concepts by themselves[1]. Thus, deep learning can capture and represent knowledge basis and reason like a real person[2].

Computer vision is a science to process the information and gain high-level understanding from digital images and videos. Once developed and perfected, it can serve as a human vision system for AI agents. The recent breakthrough achieved for the large-scale image classification on ImageNet[3] using Convolutional Neural Network (CNN) significantly accelerated the development of vision-based technologies, and deep learning became an essential tool for computer vision. Two common techniques, classification and segmentation, are used in practice for interpreting the scenes represented in the images or videos acquired from cameras. Categories of objects are predicted through image classification but, in image segmentation, the pixels are labeled by classes of the objects (i.e., semantic segmentation) or the objects are marked by masks (i.e., instance segmentation)[4].

Since structural damage captured by the cameras can be good, well-focused or not, the quality of the damage viewed from field investigations in extreme events varies and would not be at the same level. In addition to the variation of structural damage on different materials, decoration layers or covers on the structural components can also affect the appearance of the damage. Deep learning methods can effectively handle these types of uncertainties through data collection and training, and can make it viable for AI applications on SDD and Structural Health monitoring (SHM) with vision-based technologies[5]. For example, wall-climbing robots and Unmanned Aerial Vehicles (UAVs) had been used in real projects with deep learning networks for collecting and detecting the cracks on a tunnel[6]. Deep learning is the technique with great potential for measuring and assessing the damage observed in laboratory experiments, field investigations, and annual inspections of existing infrastructures[7].

Department of Civil, Environmental and Geodetic Engineering, The Ohio State University, USA.

Corresponding author:
Yongsheng Bai, Department of Civil, Environmental and Geodetic Engineering, The Ohio State University, Columbus, USA.
Email: bai.426@osu.edu

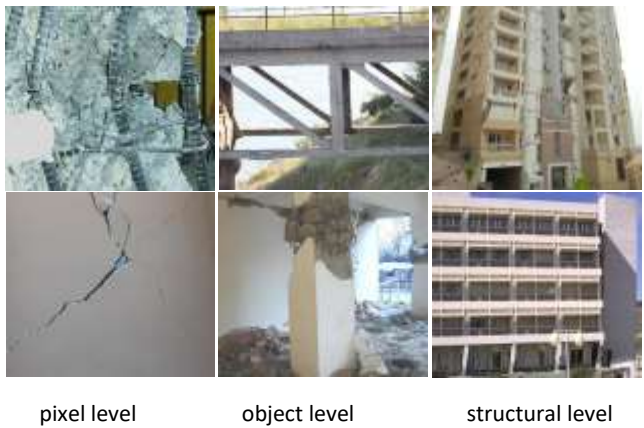

pixel level     object level     structural level

**Figure 1.** Three scene levels (scales) in our models

The observability and detectability of structural damage are affected to great extent by their scales in images. For structural damage captured at a varying scale, e.g., when cameras are closer to or farther away from them, the detected shapes and number of the damaged regions may look quite different. The appearance of background structural elements will also change in images[8]. Figure 1 shows the examples of cracking and spalling damage at three scales or three scene levels: pixel, object and structural levels. At the pixel level, cracks and spalling are clearly captured but structural components including columns, beams, walls and slabs are partially captured and cannot be identified accurately. These components can be recognized in object-level images. At the structural level, the entire structures, e.g., buildings or bridges, can be observed along with the damage. The cracks and spalling, having various shapes and depths depending on the surroundings, are less visible or even invisible at larger scale. Therefore, it is necessary to include representative images at different scales before annotations. This is the best way to counteract the imbalance of training samples in practice[9].

Aiming at practical solutions for AI robots in extreme events (e.g., earthquakes) when human experts may not be readily available in the affected regions or it may be too dangerous for engineers to closely inspect the damaged infrastructures, this research attempts to reduce the reliance on human experts by developing and improving the deep learning procedures to automatically detect and classify the infrastructure damage.

Four consecutive SDD studies are conducted with deep learning methods in our research. These studies include: 1) classification of eight classes of damage by a ResNet, 2) a pipeline with two-step networks called cascaded networks to classify and locate the damage such as cracks and spalling, and 3) a solution for detecting the damage (e.g., cracks or spalling and cracks together) directly with the state-of-the-art deep learning methods. The flow diagrams that show how these networks work for predicting structural damage are illustrated in Figures 3, 4 and 9. Based on our knowledge, very limited research have been conducted to address the difference between classification and detection of structural damage with deep neural networks[9], and no solution is provided to unify them. Also, few image datasets collected from large earthquake events have been tested to automatically detect the damage and address the feasibility of applications with the deep learning methods in these events. In addition, we applied the proposed end-to-end deep learning method to automatically detect damage that occurred in the field during the gradual collapse of a building. Our objective is to find a generalized solution for SDD on classifying multiple types and levels of damage on reinforced concrete and masonry structures and localizing the damage at various scales using the images collected from field investigations or laboratory experiments. Our studies also aim to perform real-time SDD after finalizing all the parameters and achieving stable performance with AI agents. Thus, a structural engineer can utilize UAVs or ground vehicles to quickly and safely access the structures following an extreme event or during a periodic inspection. This will reduce the workload of structural engineers and improve the efficiency of the damage assessment during field inspections. The data obtained and used in our research and the codes will be made available to public for reproducibility, general uses, and continued work.

## Related work

Our studies benefit from many prior works which can be categorized into classification and segmentation techniques with various deep learning methods. We also discuss the datasets that were used for training and testing because they are critical for successful application of these methods.

### Datasets and damage classification with deep learning in SDD and SHM

There are several important classification datasets for SDD and SHM in the research community. Yeum et al.[10] collected a large image dataset for post-event building reconnaissance and used AlexNet to classify and identify the post-event structural damages in buildings using large scale images, such as collapse classification and identification of building components. A Regional Convolutional Neural Networks (R-CNN) was employed to localize the spalling damage in some images. Furthermore, they also provided datasets about Global Positioning System (GPS) devices, structural drawings, timestamp, and measurements to automatically classify the context information when these images were documented during post-event field investigations[11]. Meanwhile, Gao and Mosalam[8] set up PEER Hub ImageNet (Phi or $\varphi$-Net) Challenge to encourage researchers to test their methods on a collection of building structural failures. There are 36,413 pairs of images and labels at various scales in this benchmark dataset. $\varphi$-Net dataset contains eight classification tasks: 1) pixel, object, and structural scene levels; 2) damaged or undamaged state of the structures; 3) spalling or non-spalling; 4) material types such as steel, concrete and others; 5) various types of collapse of the structures; 6) component types like beams, columns, walls and others; 7) damage levels or severity; and 8) damage types for cracking, including bending-related damage, shear-related damage and combined damage, or no damage (non-cracking). To overcome the insufficiency of training data, Visual Geometry Group (VGG) network and transfer learning were employed to perform classification in their study[12]. In our studies, these images for Task 1, 3 and 8 (see Table 1) are used for training and testing the proposed methods.

## Damage detection with deep learning in SDD and SHM

Classification networks don't provide the information about where the damage is in an image. Therefore, structural engineering experts need to locate the positions and identify the type of damage by themselves while it is impossible for non-professionals to do that. For example, the bottom right image in Figure 1 shows there are spalling and cracking damage on the columns in the first and second stories of the building. The classification models would associate this image to the corresponding classes of the damage but cannot give the locations of the damage. Therefore, a segmentation network, which gives the class of each object and locates it with bounding box or mark it with masks, is needed for localizing such damage in SDD missions.

Some research focused on large scale images or multiclass damage detection. Hoskere et al. conducted an experiment with 23-layer ResNet and nine-layer VGG networks to classify and segment seven classes of structural damage, including cracks, spalling, exposed reinforcement, corrosion, fatigue cracks, asphalt cracks, and no damage[13]. Ali et al. applied Faster R-CNN on defects detection in historical masonry buildings with high-resolution images[14]. Kong and Li described an application that detects and tracks the propagation of cracks in a steel girder with a video stream[15]. Atha et al. explained the different effects when they utilized two algorithms of CNNs on detecting metallic corrosion[16]. Mondal et al. used Faster R-CNN to automatically detect four common types of structural damage, including surface cracks, facade and concrete spalling, and severe damage with exposed rebars and buckled rebars. They used bounding boxes to identify the positions and boundaries of these damage[17].

Pixel-level damage detection is a popular task among researchers. Zhang et al. proposed an improved CNN for autonomous detection of pavement cracks at the pixel level[18]. Liu et al. demonstrated the application with U-Net to segment the crack on concrete structures[19], and their experiment shows that the proposed network outperforms the CNN which was used by Cha et al.[20]. Dung and Anh also used Fully Convolutional Network for localizing the cracks on the concrete surface[21], Liu et al. implemented DeepCrack, which is made of an extended Fully Convolutional Networks (FCN) and a Deeply-Supervised Nets (DSN), to pin out pixel-wise cracks[22].

Recent research became more applicable. With Holistically-Nested Edge Detection (HED) network and U-Net, Yang et al. detected cracks and spalling on concrete structures and then reconstructed 3D models through Simultaneous Localization and Mapping (SLAM) using drone images[6]. Cha et al. utilized Fast R-CNN for locating five types of structural damages, including concrete cracking, steel corrosion with two levels (medium and high), bolt corrosion, and steel delamination.

There are a total of 2,366 labeled images with the size of 500×375 for training[23]. Kim and Cho automatically localized the cracks on a concrete wall with Mask R-CNN and employed an additional image processing procedure on each bounding box to quantitatively measure the width of these cracks. The training data included 376 images[24]. Based on 1,250 images with sizes varying from 344 × 296 to 1,024 × 796, Kalfarisi et al. employed structured random forest edge detection in the region of bounding boxes of a Faster R-CNN to localize the cracks and compare it with Mask R-CNN. Photogrammetry software was used to reconstruct 3D model, thus, the cracks can be visualized and quantified further[25].

The aforementioned research help us to collect data and create training datasets when we began to our studies, to understand how to use deep learning methods correctly and effectively, and to find right solutions for the problems we are facing on in practice.

## Data preparation and methodologies

Data should be prepared for training and testing when the applications on SDD with deep learning methods have different objectives and expectations. Each method has its own requirement on the size and composition of the visual data. In general, there are more available datasets for a classification network than for a segmentation network on detecting structural damage, since the latter needs more efforts for labeling the image data[26].

### Tools for data preparation

Since the SDD models are trained in a supervised way (i.e., damage is clearly defined in training datasets), data collection and preparation are vital to train and validate the deep learning models. In current research, these data must be labelled manually by structural engineers or by people with civil engineering background. Many researchers[10,23,12,9] have shown how to perform data preparation, such as associating the classes of damage in the images, annotating the location of the damage with a bounding box, or defining the boundaries and shape of the damage for structural damage classification or detection.

In our studies, COCO (Common Objects in Context) Annotator[27] is chosen as a tool to label the damage (i.e., cracks and spalling) on infrastructure, such as buildings, bridges and other structures except for steel structures, and even on some non-structural components for training and validation. The boundaries of damage on an image are defined with polygons to form a closed region to represent them, so that the error of the damage shapes will be no greater than one pixel when labeled. Then, these regions are converted to the labels with binary images and saved as a JSON (JavaScript Object Notation) file for training. Figure 2 shows some examples of original images and labels of cracks and spalling in our training datasets. Noted that each type of damage belongs to a class and is independent of each other during labeling, although they look overlaid in some annotated images when they appear at the same location.

### Structural damage classification with a ResNet

ResNet provides higher accuracy than networks like VGG and GoogleNet because of its unique framework. The residuals in each layer of this neural network can be set to zero and the whole hierarchical feature combinations can be optimized original          label          original          label with skipping connections. Therefore, the network can be designed to have large number of layers for extracting high-level features[28]. A 152-layer ResNet with transfer learning and fine-tuning technique were used in our damage classification network for PEER Hub ImageNet (PHI) Challenge[12] where our approach secured the third place during the competition[29] (see results at https://apps.peer.berkeley.edu/phicha-llenge/winner/). The flowchart of our ResNet is shown in Figure 3.

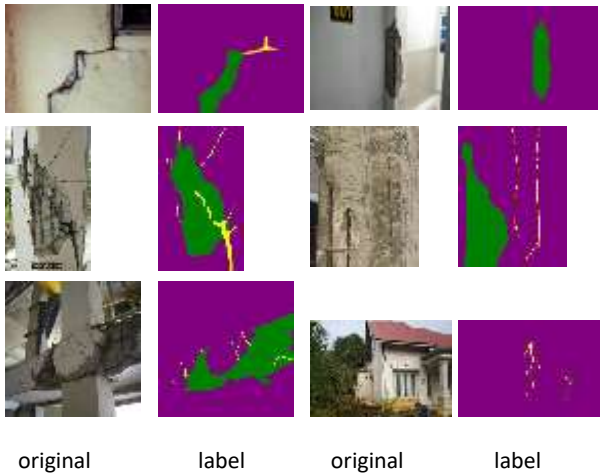

**Figure 2.** Some examples of training data (Cracks and spalling are in yellow and green while background is in purple for each labeled image).

### Cascaded networks for structural damage classification and localization

After using the ResNet to categorize various structural damage, material types and even the severity of damage in these images, we still need to delineate the damage location during SDD. Without delineation the structural engineers would still have to mark the damage locations manually. Instead of that structural engineering experts must be involved and manually identify the damage, a new pipeline similar to a process of diagnosing an illness using a combination of doctors' personal experiences, medical equipment, and available patient data is provided[9]. In this study, the classification network as a classifier can tell whether there are structural defects in the selected images, and another segmentation network (e.g., U-Net) serves as a detector to locate them. These two-step networks are named as cascaded networks. Its flowchart for detection is shown in Figure 4.

In cascaded networks, the existing classification networks used by researchers on SDD can be kept without any change, but a segmentation network is added after structural damage being categorized. From among other architectures, U-Net[30,31] is chosen as the detector to locate the damage in our study.

The U-Net has a symmetrical structure in down-sampling and up-sampling process, and each layer of down-sampling is connected to the corresponding layer of up-sampling. Thus, low-level features in the down-sampling can be directly absorbed to high-level features during the up-sampling. In our practice, the cascaded networks are utilized as a method to find the positions of cracks and spalling in images.

### Structural damage detection of cracks and spalling with Mask R-CNNs as an end-to-end method

The latest Mask R-CNNs are tested and updated for detecting the damage directly because cascaded networks are not an end-to-end method but a two-step network, which may be time-consuming and complicated for a structural engineer. In addition, high-resolution images have to be resized to low definition in the cascaded pipeline (more details are discussed in the implementation section), which affects the visibility of structural damage in images. Mask R-CNN is a benchmark of regional convolutional neural networks for instance segmentation. It is based on Faster R-CNN[32], which uses Regional Proposal Network (RPN) to automatically produce the proposals for Region of Interest (ROI) on feature maps convoluted from the original image and achieves higher speed and accuracy at low computational cost. This is also the first stage for the Mask R-CNN. In the second stage, Mask R-CNN continues to predict the damage like spalling on an image by classifying it and regressing it with a bounding box. Then a mask of the damage is created within its boundaries and shapes in the third stage for each ROI when ROI Align is utilized (see Figure 5). Furthermore, ResNet and a Feature Pyramid Network (FPN) are incorporated to obtain high quality feature maps[33].

Three variants of Mask R-CNNs are explored and developed in our studies. Meanwhile, with the awareness of the scale problem on SDD, new datasets, which include all three defined scales, are prepared and newly-developed skills in deep learning methods are also employed for improving its performance.

1) Mask R-CNN with Path Aggregation Network (PANet) and spatial attention mechanism: PANet is different from the original Mask R-CNN in its first and second stages[34]. New connections between low feature maps and high feature maps in a FPN has been introduced (see Figure 6a), which increase the efficiency of feature extraction. In addition, a technique called adaptive feature pooling is used to fuse all levels of features in a proposed ROI at stage 2 in Figure 6c. The other procedures and stages are the same as a traditional Mask R-CNN. To improve the level of feature extraction, spatial attention mechanism[35] is also introduced into this framework. We will refer this modified version as APANet Mask R-CNN in this paper.

2) Mask R-CNN with High-resolution Network (HRNet): HRNet is a state-of-the-art backbone developed for feature extraction and applies multi-scale fusion across the convolutional blocks[36]. With a traditional FPN, the features of an instance are embedded from the image via down-sampling to obtain different levels of features, or in other words, rescale the image from original size to a smaller size using convolutional operations at each level to obtain higher-level features (see Figure 6a). But HRNet has a parallel structure as shown in Figure 7. After each convolutional operation block, down-sampling process is utilized for high-level features as a FPN does. However, a new branch is also created to keep the size of feature maps at this level. By introducing low-level features with a strided convolutional operation and combining high-level features with up-sampling operations, these new branches continue to convolve until last step of final stage. Since these inter-connections between branches are used, more useful features can be extracted through this

Cascade Mask R-CNN introduces another two prediction branches with different strategies to detect and segment instances (see Figure 8b to 8d). Since more pooling operations on ROI proposals on the feature maps, it can overcome the overfitting problem existing in original Mask R-CNN[37].

These Mask R-CNNs are used to predict the class of the damage, limit the damage range with bounding boxes, and mark the shapes and boundaries of the damage with masks on a new image as the results of an output. Figure 9 shows the flowchart for this end-to-end methodology for testing on new images. All the training parameters and techniques of these deep learning methods are illustrated in the Implementation section.

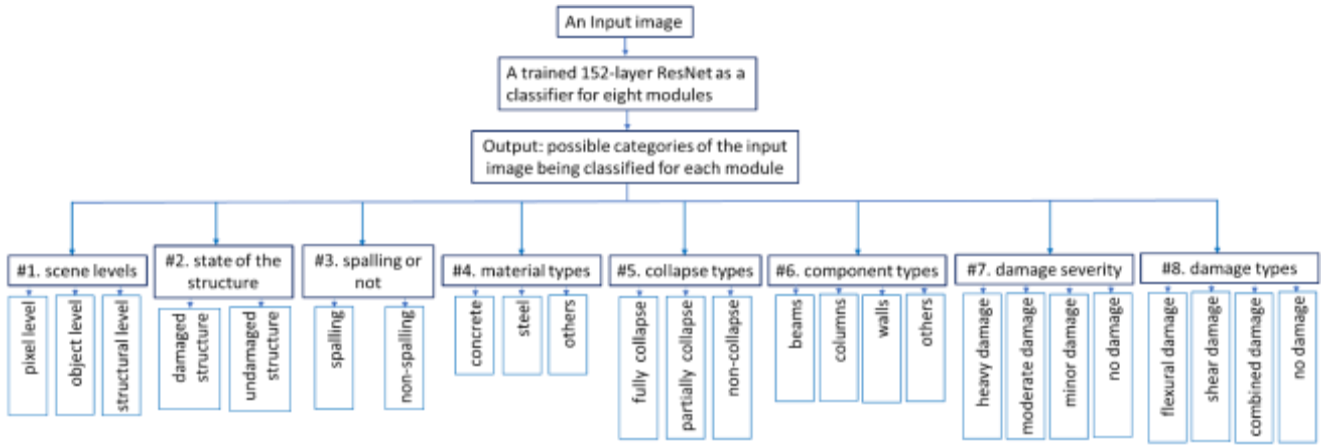

**Figure 3.** Flowchart of our 152-layer ResNet for structural damage classification.

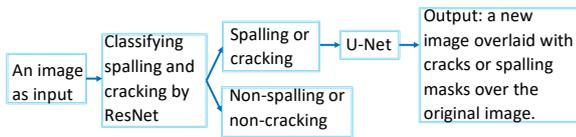

**Figure 4.** Flowchart for cascaded networks for testing.

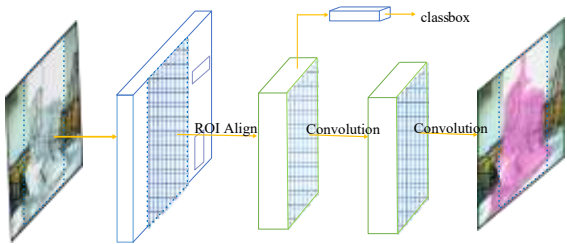

**Figure 5.** Framework for Mask R-CNN in a SDD task (a mask is in purple for the spalling damage on the column, and a bounding box is represented with dash green line).

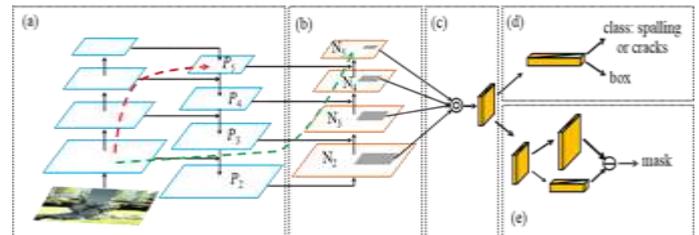

**Figure 6.** Framework for PANet in SDD for detecting cracks and spalling: (a) FPN backbone; (b) Bottom-up path augmentation; (c) Adaptive feature pooling; (d) Box branch; (e) Fully-connected fusion. $P_i$ and $N_i$ in the figures denote the $i$th of the original pyramid layers and new feature layers.

## Implementation details and results

In this section, the implementation details of the aforementioned techniques for automatic classification and detection of structural damage are presented.

### *Automated classification for eight tasks of damage detection with the ResNet*

ResNet outperformed other networks in many classification tasks, so we chose a 152-layer ResNet to identify material or damage types, structural components, spalling or non-spalling, and even the severity of structural collapse and damage in the new network for high-resolution images. After substituting the ResNet of the original Mask R-CNN with the HRNet, it forms a new network structure which we refer to as HRNet Mask R-CNN.

3) Cascade Mask R-CNN: Mask R-CNN has two prediction branches after the first stage (i.e., it is represented by convolution blocks in Figure 8a), in which the feature maps are extracted from the input image and potential ROIs are proposed by a FPN. These two branches can generate the class C, bounding box B, and mask S for each instance on the image. They are also called detection and segmentation branches.

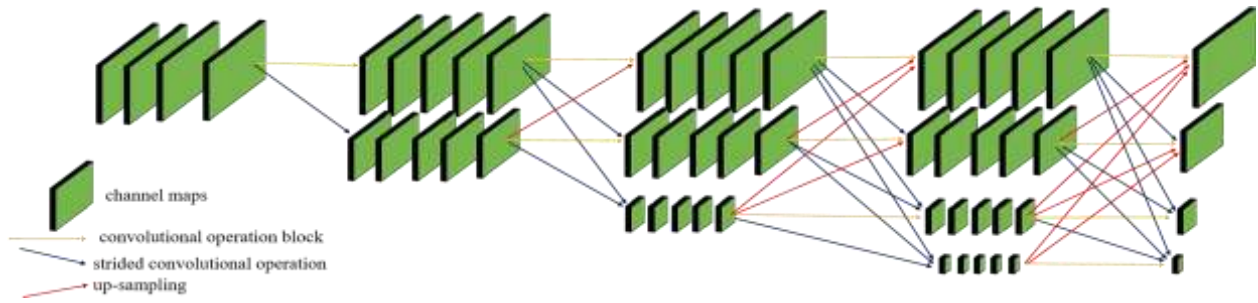

**Figure 7.** Framework for HRNet in SDD.

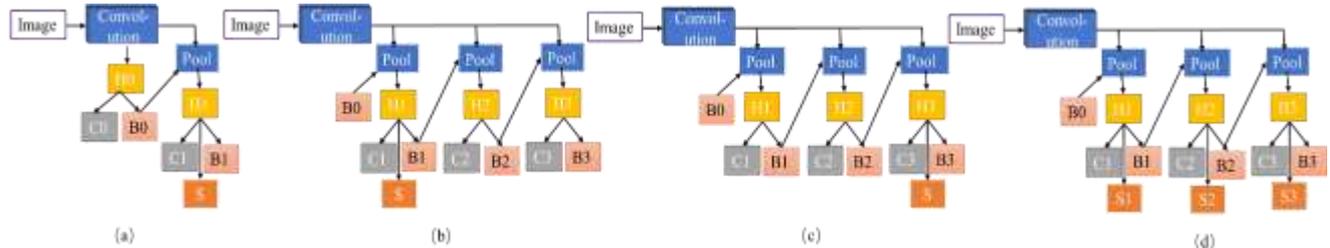

**Figure 8.** Framework for the original Mask R-CNN (a) and three Cascade Mask R-CNNs (b)-(d). "Image" is the input, "Convolution" is the backbone for convolutional operation on the input image, "Pool" is region-wise feature extraction, "H" is the RPN head, "B" is bounding box, "C" is classification, and "S" denotes a segmentation branch.

competition[12]. Image size of the $\varphi$-Net dataset is uniform as 224×224 but at various scales. For training, hyper-parameters are set as follows: learning rate is 0.001 and momentum is 0.9; the loss function is cross-entropy; 40 min-batches and 100 epochs are defined to maximize GPU usage. A NVIDIA GeForce GTX 2080 Super GPU is used for training and testing. Furthermore, the classification result of each task is evaluated by using the confusion matrix, where the diagonal elements denote true predictions. So the metric called accuracy represents the percentage of the correct performance on each task and is defined as:

$$Accuracy = \frac{(\text{Number of } True \text{ Predictions})}{N} \quad (1)$$

where $N$ = total number of samples.

Testing results are shown in Table 1. The ResNet model can identify scene levels (scales) and material types with a very high accuracy while the accuracy of the severity of collapse and damage and the types of damage is not very high but acceptable considering difficulties associated with such classification tasks.

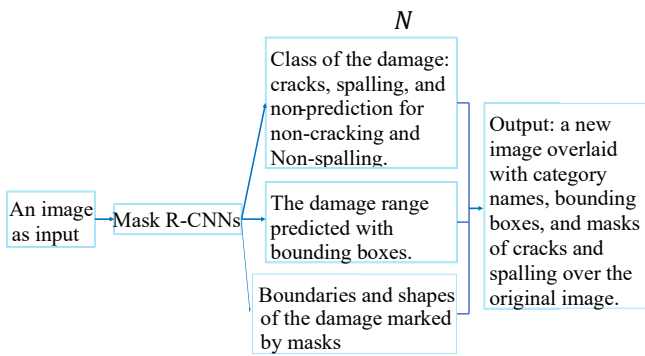

**Figure 9.** Flowchart of Mask R-CNNs used for structural damage detection in the testing stage.

## Automated damage detection with cascaded networks

Cascaded networks were used to identify and localize cracks in the first session, and then to detect the spalling damage in the second session as new datasets were introduced in this study. In the pipeline, the ResNet has the same parameters and setup as our prior study[29], while learning rate and loss function in the U-Net are set as 0.0001 and binary cross-entropy. We used the same GPU as before. In detection test with the U-Net, the accuracy is defined as the model can at least mark one piece of crack or spalling. Each pair of original image and the prediction is checked and thrown into corresponding folders of correct and incorrect predictions. thus, the ratio of the correct predictions over the total number of testing images is the accuracy of the U-Net.

1) Cascaded networks for crack detection[9]: On one hand, 1,000 images at pixel level and 853 images at object and structural levels are labeled and used for training the U-Net with COCO Annotator. All these images were resized to 256×256 in order to be compatible to the size of images in $\varphi$-Net dataset[12]. The ResNet was trained with Task 8 of $\varphi$-Net dataset (see Table 1). On the other hand, some images and another publicly available dataset called Concrete Surface Crack (CSC)[38] were tested. The latter has an image size of 227×227. Noted that all the images have no need to be resized when the U-Net was used but they were resized to 224×224 only for the ResNet in this study.

There are a total of 40,000 images in the CSC dataset, half of which are cracking and non-cracking at pixel level. We tested the proposed cascaded networks with this dataset. All the images are identified as pixel-level ones by the ResNet, and the average accuracy for cracking and non-cracking is 91.2%. For these images categorized as cracking ones, the U-Net precisely marks the cracks and does not have any failure cases in the remained 19,000 images after 1,000 images being labeled and trained.

**Table 1.** Classification results on testing data of φ-Net dataset with the 152-layer ResNet.

| Detection tasks | Number of classes | Image statistics | | | Testing accuracy of the ResNet |
|---|---|---|---|---|---|
| | | Training | Validation | Testing | |
| 1 Scene classification | 3 | 13,939 | 3,485 | 4,356 | 93.8% |
| 2 Damage check | 2 | 4,730 | 1,183 | 1,479 | 81.9% |
| 3 Spalling condition | 2 | 2,635 | 659 | 824 | 79.6% |
| 4 Material type | 2 | 3,470 | 867 | 1,085 | 99.5% |
| 5 Collapse check | 3 | 2,105 | 527 | 658 | 63.1% |
| 6 Component type | 4 | 2,104 | 526 | 658 | 71.7% |
| 7 Damage level | 4 | 2,105 | 527 | 658 | 67.8% |
| 8 Damage type | 4 | 2,105 | 527 | 658 | 67.5% |

The implementation of cascaded networks on φ-Net dataset is a dilation study. It includes these procedures in the test: First, a new testing data were selected from training and validation images in Task 1 of φ-Net dataset (see Table 1). Second, the U-Net was used to mark the cracks directly and its accuracy was calculated. Third, the cascaded networks were applied on these testing data when the ResNet and U-Net was applied to classify and locate cracks in these images. Finally, the accuracy of the cascaded networks is computed. The result of this experiment is shown in Table 2. It shows that the accuracy of the cascade networks is improved dramatically because the ResNet as the first gate to filter out some images without cracks on structural elements and the U-Net can focus on less-noised images to mark the cracks.

2) Cascaded networks for spalling detection: For this test, the RestNet was trained in Task 3 of φ-Net dataset (see Table 1), and 1,178 images were prepared for training the U-Net to mark the spalling. Training data for the latter are from the collection of Yang et al.[6] and our own work[39]. There are two datasets for testing. The first one is a spalling dataset by Yeum et al.[10], in which there are 1,000 images with a uniform size of 640×480. All these images have the spalling damage shown on the structural components. When the U-Net was used, it acquired an accuracy of 99.0% on detecting the spalling. But the accuracy of the ResNet is 85.6% for classifying these images after they were resized to 224×224. On the other hand, a total of 1,692 images labeled as the spalling ones in Task 3 of φ-Net dataset were directly tested by the U-Net and its accuracy reached to 97.6%. The ResNet for φ-Net spalling dataset has an accuracy of 79.6% as shown in Table 1. We didn't use the U-Net to locate the spalling damage right after the ResNet because the former has such a high accuracy on these two datasets. The accuracy of cascaded networks for this test is determined rather by the ResNet than by the U-Net.

Some examples for good predictions from the cascaded networks are shown in Figure 10. It can be seen that the proposed networks typically provide precise locations of these cracks and spalling, so there is less reliance on human experts to manually find them. It should be pointed out that it takes more time for cascaded networks to classify and locate the damage than the ResNet or the U-net is used alone.

### An end-to-end method to automatically detect cracks and spalling by using Mask R-CNNs

Since two networks are involved in the cascaded networks and the training image data, especially those for the ResNet, have a low resolution, we realized that there is a great need to simplify the framework and make it applicable to images at different definitions. On one hand, as the high-resolution images are resized to smaller ones, the visibility of tiny damage like cracks may be reduced or even be invisible. The useful information about the damage can be saved if the size of these images has not been changed, thus, the networks can extract more precise features from original images. On the other hand, the U-Net is not suitable for segmenting multiple types of damage simultaneously. Therefore, our next goal is to find and assess end-to-end neural networks to localize the damage like cracks and spalling when they are captured at different scales and varied resolutions.

The accuracy of these predictions is redefined as these models correctly detected one or two kinds of structural damage in Eq. (1). To be more specific, it is a correct prediction when at least one piece of cracks or spalling has been marked by a bounding box or a mask when the damage is visibly captured on each testing image. Otherwise, it is also correct when no prediction is given for those images without any cracks or spalling on the structure. Each prediction from the models and the original images are compared before being moved into the corresponding folders of correct or incorrect detection. In addition, parameters are set for the training of Mask R-CNNs as follows: learning rate, momentum and decay rate of weights are defined as 0.002, 0.9 and 0.0001, respectively; the loss function for the mask is cross-entropy and for bounding boxes is smooth L1; the number of epochs is 100 with the same GPU used in our previous studies. Our source codes are originated from MMDetection[40] while some modifications and updates were made for different purposes during the training and testing.

1) Crack detection with APANet Mask R-CNN and HRNet Mask R-CNN[26]: This pipeline is an end-to-end solution for damage detection, in which a total of 2,021 images with the size from 168×300 to 4,600×3,070 and with three scales were labeled. The annotated data format is similar to the examples shown in Figure 2 while the original images came from our own collection and internet search. The performance of APANet Mask R-CNN on φ-Net dataset is shown in Table 2. Compared to the cascaded networks, this Mask R-CNN has a dramatic improvement for these images at larger scale other than those at pixel level. In addition, APANet and HRNet Mask R-CNNs were directly applied onto 2017 Pohang earthquake images dataset (PEI2017)[41] and 2017 Mexico City earthquake images dataset (MCEI2017)[42], which include 4,109 and 4,136 high-resolution images collected by structural experts after these two Richter magnitude 5.2 and 7.1 earthquakes happened. Table 3 shows the accuracy of two models, but both Mask R-CNNs have a close accuracy on two testing datasets. This test indicates that it is possible for end-to-end deep learning methods like the latest Mask R-CNNs to precisely detect cracks at various scale in large earthquake events.

**Table 2.** The accuracy of the U-Net, cascaded networks and APANet Mask R-CNN (APAN Mask) on detecting cracks in $\varphi$-Net dataset (TP means true predictions).

| Scene levels | Number of images | U-Net | | Cascaded networks | | | APAN Mask | |
|---|---|---|---|---|---|---|---|---|
| | | TP | Accuracy | ResNet | U-Net | Accuracy | TP | Accuracy |
| Pixel level | 4,661 | 2,819 | 60.5% | 2,988 | 2,810 | 94.0% | 3,948 | 84.7% |
| Object level | 5,713 | 1,490 | 26.2% | 1,479 | 1,129 | 59.6% | 4,407 | 77.1% |
| Structural level | 5,832 | 500 | 8.6% | 717 | 356 | 49.7% | 4,774 | 81.9% |

2) Spalling and crack detection with new variants of Mask R-CNN as an end-to-end method[39]: Since APNet and HRNet Mask R-CNNs worked quite well for crack detection, we added the spalling damage into the detection task to check whether this solution is more robust and applicable in field investigation. Image collection from Yang's spalling dataset[6] and our in-house generated dataset were relabeled, resulting in a total of 2,229 curated images for training and validation.

**Table 3.** Accuracy of APANet and HRNet Mask R-CNN for crack detection on two public datasets.

| Methods | PEI2017 | MCEI2017 |
|---|---|---|
| APANet Mask R-CNN | 74.1% | 70.6% |
| HRNet Mask R-CNN | 74.0% | 73.0% |

**Table 4.** Accuracy of three Mask R-CNN for cracks and spalling detection on three public datasets.

| Methods | $\varphi$-Net CrSp | PEI2017 | MCEI2017 |
|---|---|---|---|
| Cascade Mask R-CNN | 78.9% | 66.0% | 69.4% |
| APANet Mask R-CNN | 81.1% | 67.6% | 74.7% |
| HRNet Mask R-CNN | 58.6% | 68.1% | 69.1% |

Also, size of these images varies from 147×288 to 4600×3070. The examples are shown in Figure 2. In addition, a new variant of Mask R-CNN was introduced for comparison resulting in three different trained Mask R-CNNs that are tested. Data augmentation[43] was utilized for improving performance.

More diverse testing data are included for the purpose of detecting two major types of damage in extreme events. On one hand, all of the spalling, non-spalling, cracking, and noncracking images of training and validation data in Task 3 and 8 of $\varphi$-Net dataset were collected and combined into a new testing dataset, which is called $\varphi$-Net CrSp dataset in Table 4. It consists of 5,853 images in total and is a low-resolution but comprehensive dataset. On the other hand, high-resolution images from PEI2017 dataset and MCEI2017 datasets were still used for testing these three networks. The performance of these three networks on these datasets is shown in Table 4. APANet Mask R-CNN has a better performance than the other two on detecting cracks and spalling no matter what the images are at low or high resolution. Overall, these Mask R-CNNs can achieve an accuracy above 66.0% for detecting two major structural damage with high-definition images, although the performance of APANet and HRNet Mask CNNs declines a little for spalling and crack detection compared to its performance on crack detection for two same datasets.

Figure 11 illustrates the predictions made from these Mask R-CNNs on the testing datasets. Cracks and spalling at various scales and different resolutions can be identified and located automatically that will reduce the need for human interaction on locating damage following an extreme event. It should be noted that the limitation of insufficient training data at hand is the major reason for incorrect predictions in our studies. In addition, the imbalance among the datasets training for damage localization is still hard to overcome. So, the deep learning models are easily distracted by these objects which appear like cracks and spalling in images such as: 1) wires or cables; 2) trees; 3) fences; 4) shadow; 5) edges of windows, buildings or other artifact objects[39]. Through refining a few parameters in the testing step, we also observed that the exact shapes and numbers of the damage can be precisely marked in some images. For some SDD tasks, new damage-related images need to be introduced into the training datasets of this end-to-end method for better performance.

**Table 5.** Accuracy of APANet Mask R-CNN for damage detection from images collected during progressive collapse testing of a building in the field.

| Image source | Total number | True predictions | Accuracy | Average accuracy |
|---|---|---|---|---|
| Cell phone | 220 | 172 | 78.8% | |
| Drones | 303 | 172 | 56.8% | 65.8% |

## An application of damage detection in a building during a collapse experiment in the field

Recently we conducted a field experiment to study progressive collapse performance of a six-story parking garage structure on the main campus of The Ohio State University in Columbus, Ohio. The APANet Mask RCNN was used to detect

1) CSC dataset

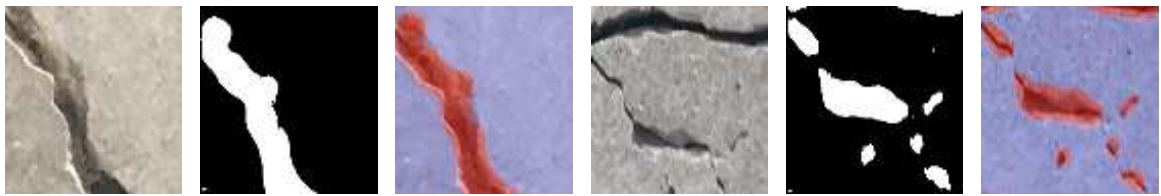

2) Pixel-level $\varphi$-Net dataset

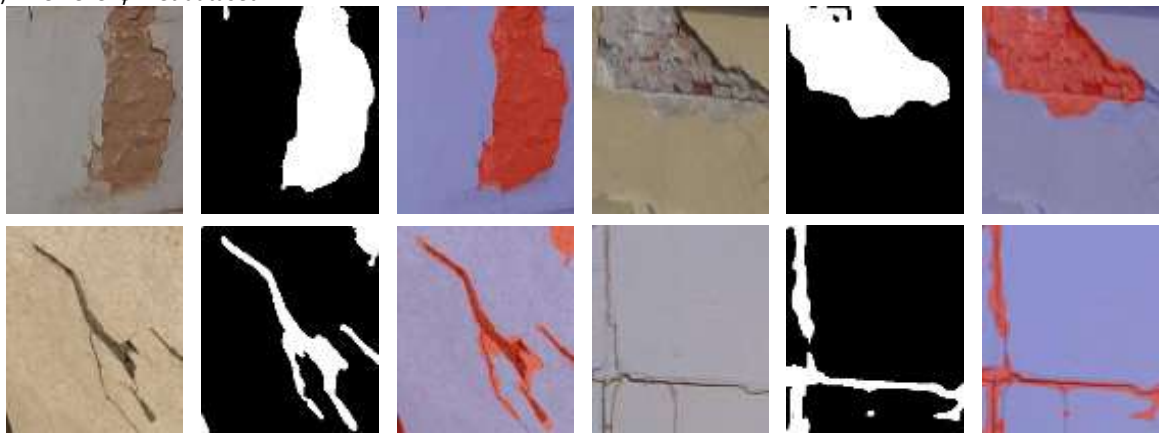

3) Object-level $\varphi$-Net dataset

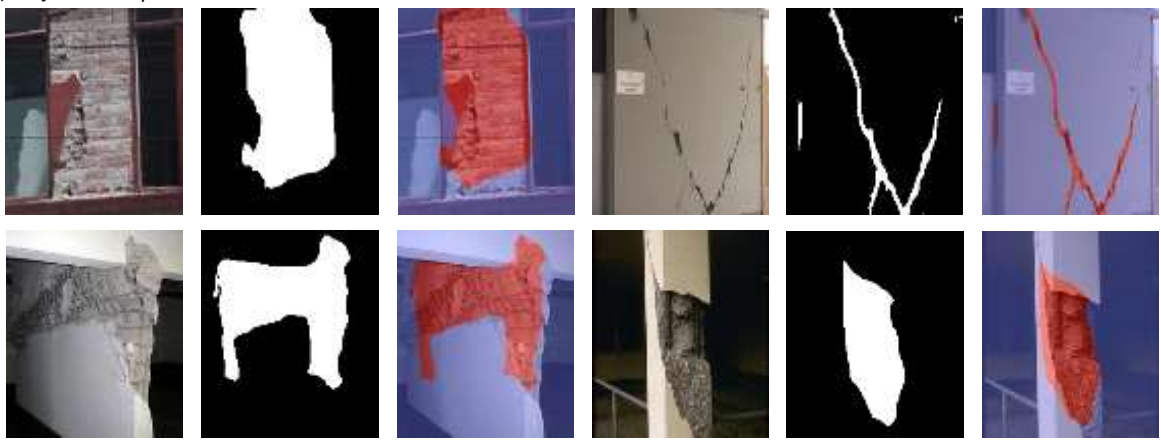

4) Structural-level $\varphi$-Net dataset

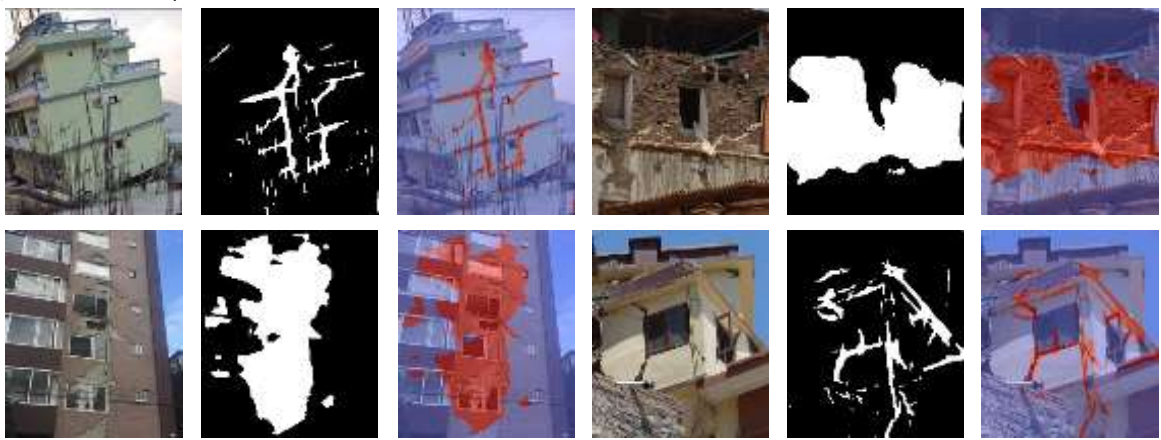

original        predicted        overlaid        original        predicted        overlaid

**Figure 10.** Some examples of correct prediction for cracks and spalling by cascaded networks (cracks and spalling are in red and white for the overlaid and predicted images, respectively, and background of the predicted images is in black).

1) φ-Net dataset

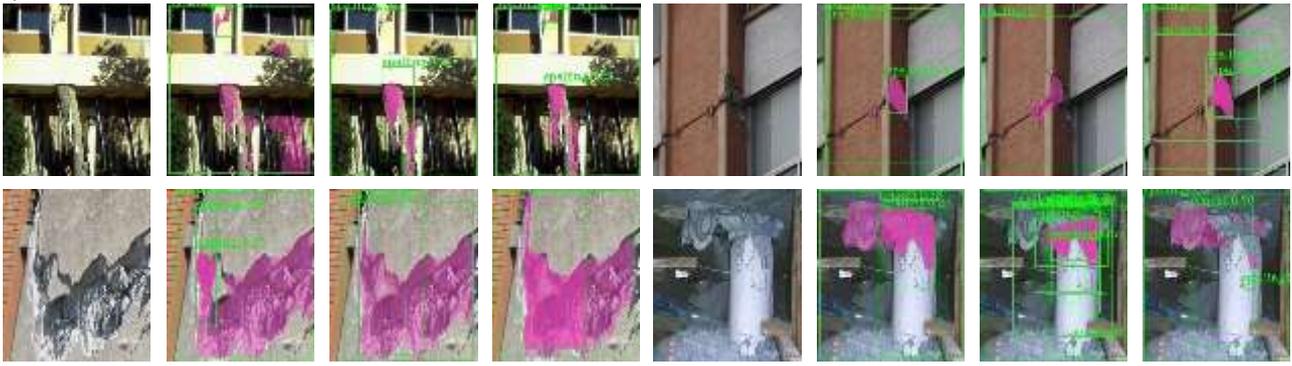

2) MCEI2017 dataset

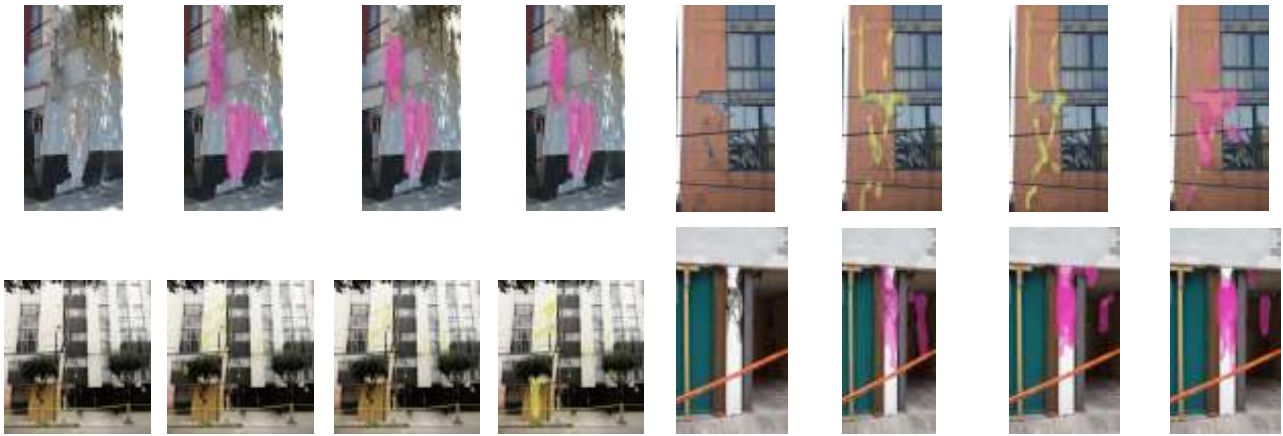

3) PEI2017 dataset

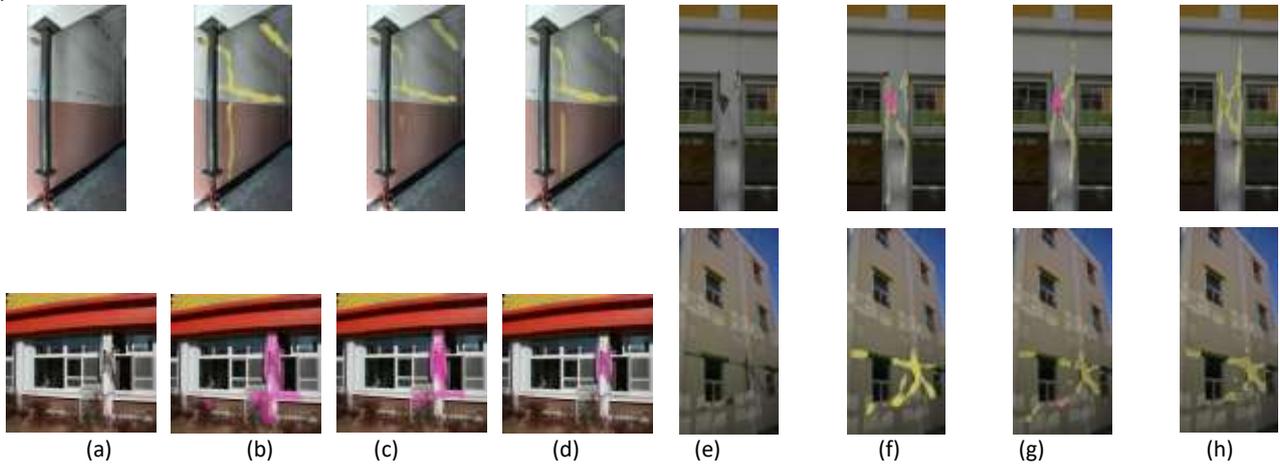

(a)      (b)      (c)      (d)      (e)      (f)      (g)      (h)

**Figure 11.** Some examples of predictions from three Mask R-CNNs for three public datasets. (a) and (e), (b) and (f), (c) and (g), and (d) and (h) denote original image, overlaid image of Cascade, APANet and HRNet Mask R-CNN, respectively. Bounding boxes, cracks and spalling are in green, yellow and purple, respectively.

the damage from visual data collected from our cell phones (iPhone 12 Pro) and drones (Wingtra and DJI). In this field experiment, the building was instrumented and portions of concrete slabs and facades at each floor level and reinforced concrete columns in the top two stories were removed from the building. Our stationary cameras inside and outside the building, cell phone cameras and drone cameras well documented the entire process of slab, facade and column removals and associated damage. Damage progression was captured from different angles at different scales during the removal of the reinforced concrete building.

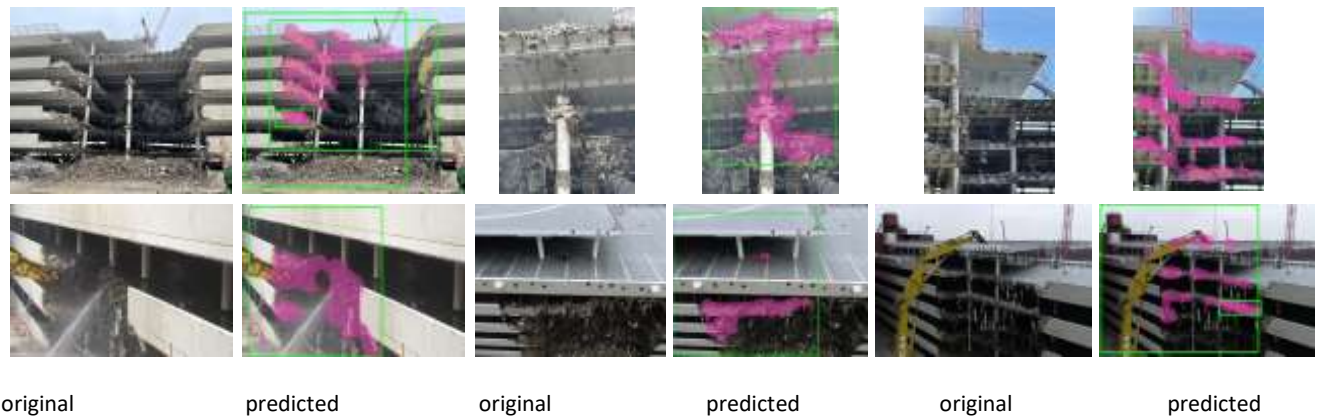

| original | predicted | original | predicted | original | predicted |

**Figure 12.** Some examples of predictions by APANet Mask R-CNN for images from a progressive collapse experiment of a building in the field. Damage is indicated by green bounding boxes (i.e., boundaries of detected damage), purple (i.e.,spalling) and yellow

For damage detection from images collected during this field experiment, we used a total of 523 images from cell phones and drones as the dataset for testing and evaluating the performance of APANet Mask R-CNN, since these images capture different damage levels (cracking, spalling, and mostly complete loss of concrete pieces) at various scales and different definitions. These images include three resolutions such as 1920×1080, 4030×3020 and 5470× 3640. Table 5 shows that the APANet Mask R-CNN works well to locate the damage from cell phone images, but a weak detection rate is observed for drone images. This is mainly because there are less images from drones in the training data. Some examples of good predictions from these images are shown in Figure 12, in which images from the cell phones and drones are in the first and second rows, respectively. In this preliminary analysis, the APANet Mask R-CNN shows its robustness in locating the overall damage in the building.

## Conclusions and future work

Several deep learning pipelines have been proposed as solutions for the classification and detection of different structural damage at various scales and resolutions in images collected after extreme events, such as large earthquakes. This research aims to improve the understanding of deep learning techniques and to make them practical and suitable for applications of automated structural damage detection. Our conclusions are as follows:

1) Our research shows that a 152-layer ResNet classifier can perform well for multi-class damage classification when transfer learning and parameter fine-tuning are utilized.

2) In addition to classifying damage, cascaded networks are used to localize the damage. In our pipeline, we added a U-Net segmentation network after the existing classification networks to achieve this. Our tests show that the cascaded networks outperform the U-Net as the only network for detecting cracks and spalling.

3) An approach for damage detection with end-to-end networks is developed with the state-of-the-art Mask R-CNNs. This approach is tested on public post-event image datasets after two new training datasets were prepared for two separate studies. With a new feature pyramid network[34] and spatial attention mechanism[35], APANet Mask R-CNN is shown to outperform the cascaded networks and the U-Net for crack detection in $\varphi$-Net dataset[12]. In the test of spalling and crack detection, it also achieves an accuracy above 67.6% for all images collected in extreme events at various scales and different definitions, and reaches an accuracy of approximately 81% for an image dataset at low resolution but with damage at various scales, which is more challenging for structural damage detection.

4) We applied the APANet Mask R-CNN for damage detection from images collected during a field experiment investigating progressive collapse of a building to study the effectiveness of this method.

These solutions with deep learning methods can not only solve problems of classification and localization of different structural damage in extreme events, but also can be used in structural health monitoring for in-service structures and automatic quantification of the damage in laboratory experiments as Woods et al.[7] did.

In the future, more images will be collected and labeled to counteract the distractions by other objects in images and to overcome the imbalance problem observed in some training data. We plan to test more image datasets collected from extreme events and quantify the damage with our methods and photogrammetry skills and conduct experiments for real-time SDD during field inspections.

### Acknowledgements

This study is partially based upon work supported by the U.S. National Science Foundation under grant no. 2036193.